\documentclass[letterpaper, 10 pt, conference]{ieeeconf}  

\IEEEoverridecommandlockouts                              

\usepackage{cite}
\usepackage{amsmath,amssymb,amsfonts}
\usepackage{hyperref}
\usepackage{algorithmic}
\usepackage{graphicx}
\usepackage{textcomp}
\usepackage{xcolor}
\usepackage{tikz}
\usepackage{lipsum}
\usepackage{multirow}
\usepackage{colortbl}
\usepackage{booktabs}
\usepackage[all]{nowidow}

\title{\LARGE \bf
Real-time Photorealistic Mapping for Situational Awareness in Robot Teleoperation
}

\author{Ian Page$^{1,3}$, Pierre Susbielle$^{1}$, Olivier Aycard$^{1}$ and Pierre-Brice Wieber$^{2}$%
\thanks{$^{1}$I. Page, P. Susbielle, and O. Aycard are with Grenoble Image Parole et Signal (GIPSA-lab - UMR Grenoble INP CNRS 5216), 38400 Saint-Martin-d'Hères, France. Emails: \{ian.page, pierre.susbielle, olivier.aycard\}@grenoble-inp.fr}%
\thanks{$^{2}$P.-B. Wieber is with INRIA. Email: pierre-brice.wieber@inria.fr}%
\thanks{$^{3}$I. Page is with Framatome. Email: ian.page@framatome.com}%
}

\begin{document}

\maketitle
\thispagestyle{empty}
\pagestyle{empty}

\begin{abstract}

Achieving efficient remote teleoperation is particularly challenging in unknown environments, as the teleoperator must rapidly build an understanding of the site’s layout. Online 3D mapping is a proven strategy to tackle this challenge, as it enables the teleoperator to progressively explore the site from multiple perspectives. However, traditional online map-based teleoperation systems struggle to generate visually accurate 3D maps in real-time due to the high computational cost involved, leading to poor teleoperation performances. In this work, we propose a solution to improve teleoperation efficiency in unknown environments. Our approach proposes a novel, modular and efficient GPU-based integration between recent advancement in gaussian splatting SLAM and existing online map-based teleoperation systems. We compare the proposed solution against state-of-the-art teleoperation systems and validate its performances through real-world experiments using an aerial vehicle. The results show significant improvements in decision-making speed and more accurate interaction with the environment, leading to greater teleoperation efficiency. In doing so, our system enhances remote teleoperation by seamlessly integrating photorealistic mapping generation with real-time performances, enabling effective teleoperation in unfamiliar environments.

\end{abstract}

\begin{flushleft}
  \small{Video:}
  {\scriptsize\url{https://www.youtube.com/watch?v=-Md49rKkV8I}}\\
  \small{Code:}
  {\scriptsize\url{https://github.com/ian-pge/GS_SLAM_teleoperation.git}}
\end{flushleft}
\section{Introduction}
\label{Sec:Introduction}

Remote exploration tasks, whether in space, deep oceans, or hazardous terrestrial settings, demand robots that can operate reliably in unpredictable and unstructured environments. Despite remarkable advances in robotics and artificial intelligence, fully autonomous robots struggle with complex terrain, sensor limitations, and real-time decisions, leaving remote missions unreliable. 

Robot teleoperation has emerged as a promising solution to these challenges by leveraging the strengths of human perception and decision-making. In a teleoperation system, a human operator, called a teleoperator, remotely operates a robot, thereby directly compensating for the limitations of onboard autonomous systems. This approach allows teleoperators to explore and perform tasks in hazardous or inaccessible sites without being physically present. 

\begin{figure}[h]
    \centering
    \includegraphics[width=\linewidth]{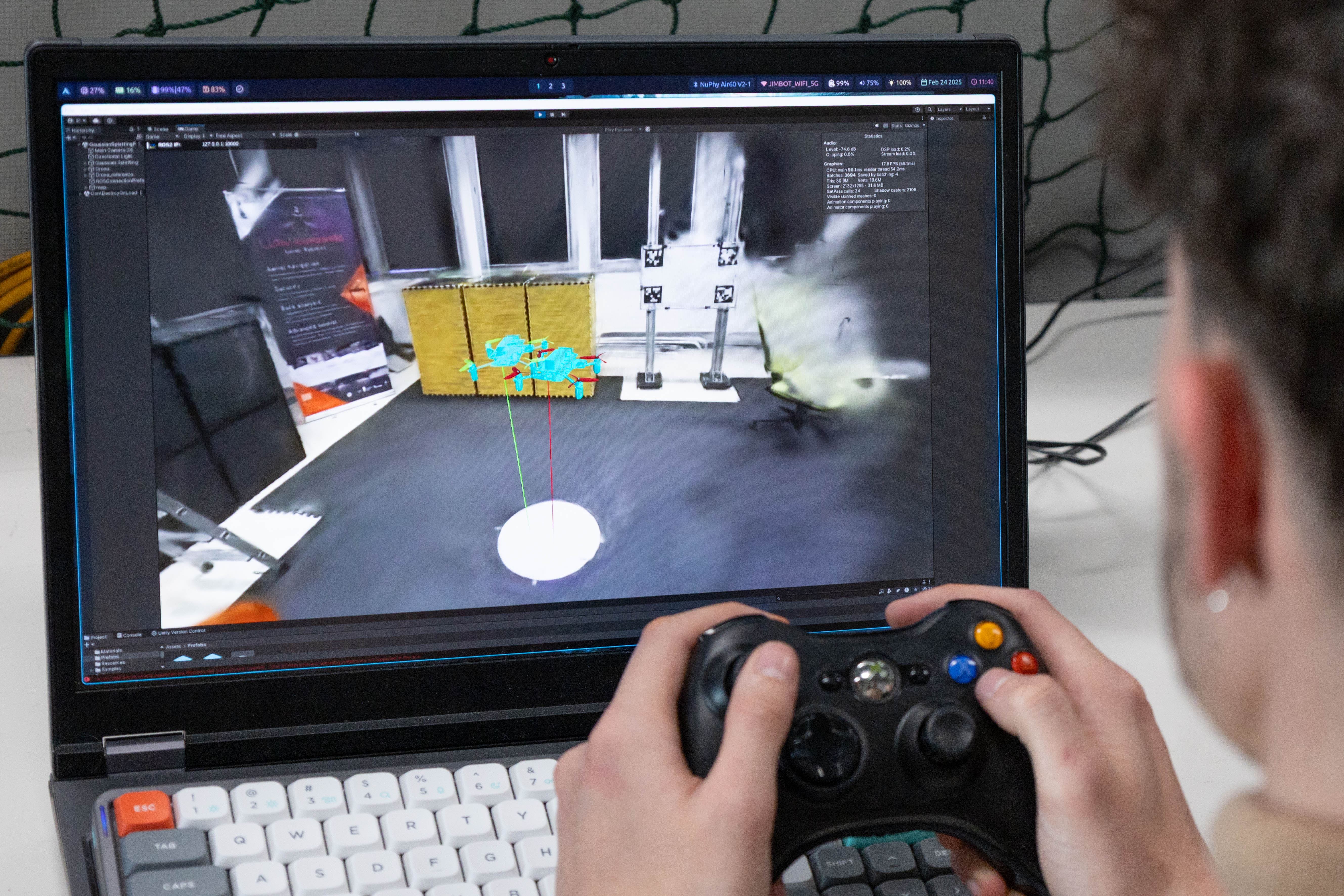}
    \caption{Teleoperator controlling an aerial vehicle at an unknown site guided by a real-time generated photorealistic 3D map. The environment is continuously reconstructed leveraging gaussian splatting SLAM as the teleoperator discovers and explores the site.}
    \label{fig:Overview}
\end{figure}

Situational awareness (SA) as stated in~\cite{endsley_toward_nodate}, is “the perception of the elements in the environment within a volume of time and space, the comprehension of their meaning, and the projection of their status in the near future”. In this work, we focus on a scenario where the site is entirely unknown, with no prior knowledge available. This contrasts with standard teleoperation, which typically takes place in a previously known environment. Achieving high SA in such a scenario is challenging~\cite{opiyo_review_2021} because remote teleoperators discover the site's layout and potential obstacles during teleoperation. SA is essential for operational efficiency, as it enables the teleoperator to interpret the environment, understand the robot's surroundings, anticipate challenges, make informed decisions, and navigate complex spaces effectively.

Traditional teleoperation systems rely on real-time 2D or 3D video feeds~\cite{lenz_nimbro_2023}, emphasizing telepresence~\cite{youssef_telepresence_2023}, which creates the illusion of being physically on-site. While this enhances immersion, it can also compromise SA during teleoperation in unknown environments. Sensory feedback may overshadow critical spatial and contextual information needed for effective navigation and decision-making. The limited SA can be attributed to:

\begin{figure*}[ht]
    \centering
    \includegraphics[width=\linewidth]{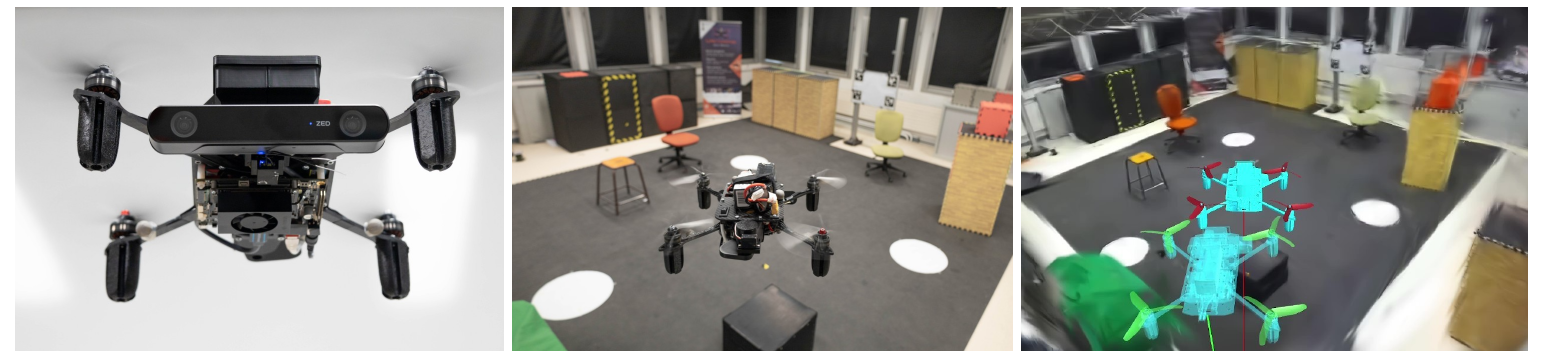}
    \caption{Overview of the teleoperation scenario. On the left, the aerial vehicle serving as the mobile robot; in the middle, the remotely teleoperated robot navigating the site; and on the right the real-time generated 3D map of the unknown environment, displaying the robot's digital twins at its current and target localizations within it.}
    \label{fig:Unity}
\end{figure*}

\begin{itemize}
    \setlength{\itemsep}{5pt}
    
    \item Limited Visual Perspective and Incomplete Mental Model: Teleoperators rely on narrow views provided by mounted cameras on the robot. This results in incomplete or inaccurate mental maps of the site, impeding scene understanding. Moreover, changing perspectives requires physically repositioning the robot, which can be both time-consuming and challenging.
    
    \item Compromised Proprioception: The teleoperator lacks sufficient feedback about the robot's current state, including position, orientation, and movement relative to its close environment making it difficult for the teleoperator to accurately perceive and control the robot.
    
\end{itemize}

A promising solution, is to provide the remote teleoperator with an interactive viewer displaying a 3D map of the site built online via Simultaneous Localization And Mapping (SLAM)~\cite{stotko_vr_2019}. This helps teleoperators build an accurate mental model, explore from any viewpoint, and maintain proprioception by visualizing the robot’s position relative to its surroundings. Even with communication delays, they can continue operating using the current map.

In this work, such a system is referred to as an online map-based teleoperation system. We assume that the environment is static, meaning that once a region has been mapped, it does not change over time. This assumption, like in all other projects discussed in this paper, simplifies map consistency and ensures that previously reconstructed areas remain reliable for navigation.

In an online map-based teleoperation system, the teleoperator's level of SA can be evaluated based on the performances of the interactive viewer, as it serves as the sole point of interaction between the teleoperator and the robot. Its performances can be evaluated based on the following criteria:

\begin{itemize}
    \setlength{\itemsep}{5pt}

    \item Map accuracy: Is the map visually and geometrically reliable?
    
    \item Map generation speed: How quickly is the map generated?    

    \item System responsiveness: Is the interaction fluid, without noticeable lag or delay?

\end{itemize}

The ideal online map-based teleoperation system, designed to maximize SA, would provide the teleoperator with a photorealistic map created instantly as teleoperation begins. However, generating visually highly accurate 3D maps in real-time is computationally intensive and introduces latency in the map creation process. These two requirements, are inherently at odds, making it difficult to optimize both simultaneously. As a result, the teleoperator must choose between map quality and system usability.

In this work, we introduce the first teleoperation system capable of delivering novel photorealistic views of unknown sites while sustaining real-time map generation and interaction, thereby eliminating the long-standing trade-off in online map-based teleoperation system. Our approach builds on established teleoperation concepts and recent research that integrates Gaussian Splatting (GS) \cite{kerbl20233d} into classical teleoperation systems \cite{wilder-smith_radiance_2024}. GS is a technique that encodes scenes using anisotropic 3D gaussians instead of dense meshes or point clouds. GS enables the creation of photorealistic maps in reasonable time, enabling highly visual accurate teleoperation in previously known environments.

The main contributions of this study are threefold:
\begin{itemize}
    \setlength{\itemsep}{5pt}
    
    \item A solution to tackle the lack of SA during teleoperation
    in unknown environments based on GS-SLAM. GS-SLAM is an innovative approach that combines the high-fidelity rendering capabilities of GS with the real-time environmental mapping and localization strengths of SLAM.
    
    \item A-ready-to-use implementation of the proposed teleoperation system, including a novel modular and efficient GPU-based integration between GS-SLAM and an interactive viewer, ensuring high visual accuracy, real-time map generation, and seamless system responsiveness without compromise.
    
    \item A comprehensive comparison of the proposed teleoperation system’s performance against existing solutions, followed by an in-depth evaluation in real-world scenarios, demonstrating its superiority over state-of-the-art approaches.
    
\end{itemize}

To the best of our knowledge, this work is one of the first real-world applications of GS-SLAM.

The paper is structured as follow. An extensive literature review on map-based teleoperation systems, along with their current limitations, followed by a discussion on GS-SLAM are presented in Sec.~\ref{sec:related_works}. The methodology for building and implementing the proposed solution is described in Sec.~\ref{sec:methodology}, and its performances are compared to existing works in Sec.~\ref{sec:teleop_performances}. Finally, real-world experiments are conducted to validate the proposed system, as detailed in Sec.~\ref{sec:experiments}. The experimental protocol includes remote exploration activities and interaction with the unknown site using an aerial vehicle.

\section{Related works}
\label{sec:related_works}

\subsection{Map-based teleoperation systems}

Map-based teleoperation systems often rely on precomputed maps~\cite{conte_design_2020}, however when exploring unknown environments, real-time mapping is required, often leveraging SLAM methods~\cite{alsadik_simultaneous_2021}. Photorealistic maps enhance visual fidelity but require significant computational resources, creating a trade-off with real-time generation~\cite{stotko_slamcast_2019}, which is typically prioritized in existing online map-based teleoperation systems~\cite{conte_design_2020, stedman_vrtab-map_2022, kuo_development_2021, kim_mapping_nodate, du_human-guided_2016, xiao_three-dimensional_2020, xie_generative_2021}.

Gaussian Splatting (GS)~\cite{kerbl20233d} is a recent breakthrough in fast photorealistic 3D scene reconstruction, originally developed as a faster alternative to neural radiance fields~\cite{mildenhall_nerf_2020}. GS represents a scene explicitly using 3D Gaussians defined by their mean, covariance and opacity, modeling geometry and view-dependent colors via spherical harmonics. GS surpasses other photorealistic methods in achieving higher visual fidelity with lower computation times, making it ideal for map-based teleoperation systems~\cite{wilder-smith_radiance_2024, bowser_3d_nodate}. GS is fundamentally a precomputed map because it requires two pre-processing steps before use. First, an initialization phase reconstructs geometry from a sparse point cloud and estimates camera poses. Then, a training phase optimizes the Gaussian parameters using a GPU-accelerated rendering algorithm.

Different approaches have been carried out to try to incorporate GS into online map-based teleoperation systems. In~\cite{li_reality_2024}, Real-time sensor data is overlaid on a GS map to provide live updates to the teleoperator. However, the GS map serves as a static reference and remains precomputed. The study in~\cite{wilder-smith_radiance_2024} introduces the first teleoperation system that generates a GS map online by incorporating SLAM elements into GS algorithms, effectively removing the need for GS pre-processing steps. This successfully combines GS's photorealism with SLAM's online map generation. However, its SLAM integration only adapts the GS algorithm without fundamentally redesigning it, allowing online map generation at reasonable speeds but not in real-time, making it unsuitable for our use-case.

Despite recent advancements, the current state-of-the-art in online map-based teleoperation systems still faces a trade-off between achieving high visual accuracy and real-time map generation. 

\subsection{GS-SLAM}

\begin{figure}[h]
    \centering
    \includegraphics[width=\columnwidth]{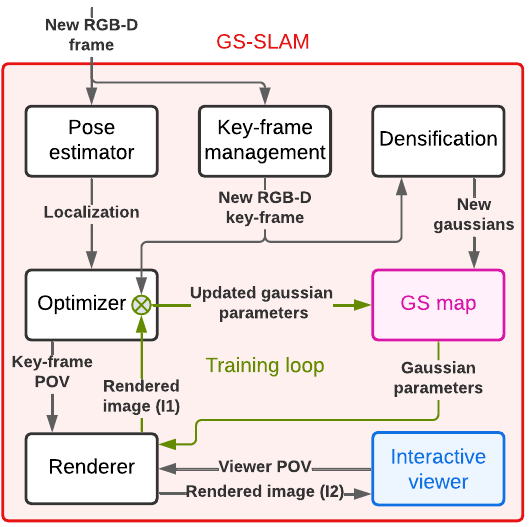}
    \caption{Generic overview of GS-SLAM. Each new RGB-D frame is first processed by a camera pose estimator. A key-frame management module decides when new key-frames should be added to densify the GS map by introducing new gaussians. A renderer utilizes the current GS map to simultaneously perform two rendering tasks: (I1) guiding the gradient-based optimization of gaussian parameters in a continuous training loop (indicated by green arrows) by comparing generated images with original key-frames; and (I2) updating views in the interactive viewer based on viewer Point of View (POV) requests.}
    \label{fig:GS_SLAM}
\end{figure}

By deeply modifying GS, recent research has overcome the challenge to integrate GS with SLAM, as shown in Fig.~\ref{fig:GS_SLAM}, resulting in the development of GS-SLAM~\cite{tosi_how_2024}. This approach involves certain trade-offs, such as simplifying the gaussian representation and eliminating view-dependent color reflections~\cite{tosi_how_2024, matsuki_gaussian_nodate}.

One approach to GS-SLAM, integrates GS scene reconstruction and localization in a single step, using a unique GS map, making it straightforward to implement~\cite{matsuki_gaussian_nodate, keetha_splatam_2023, peng_rtg-slam_2024, sun_mm3dgs_2024, sun_high-fidelity_2024, yan_gs-slam_2023, yugay_gaussian-slam_2023, wang_endogslam_2024, hu_cg-slam_2024, sandstrom_splat-slam_2024, deng_compact_2024}. However, this approach is computationally intensive, as it requires repeatedly rendering the scene and optimizing the pose to align the observed images with the GS map for camera position estimation. 

To mitigate this computational cost, an alternative approach involves separating GS scene reconstruction and localization into two independent threads. This generally requires maintaining a GS map for scene reconstruction and a separate feature map used for localization~\cite{lang_gaussian-lic_2024, jiang_tambridge_2024, li_ngm-slam_2024, jiang_3dgs-reloc_2024}. 

Recent advancements in GS-SLAM methods enable the best of both worlds: Sharing only a single map while running GS scene reconstruction and localization independently. To this extend, GS-ICP, shares covariance information between mapping and localization threads, using a Generalized Iterative Closest Point method for localization~\cite{segal_generalized-icp_2009, ha_rgbd_2024}. Similarly, Photo-SLAM leverages ORB-SLAM3 for localization while maintaining a single hyper primitive map that encodes both geometric features and photometric information, enabling loop closure and improving localization accuracy~\cite{campos_orb-slam3_2020, huang_photo-slam_2023, ho_detecting_2007}. CaRtGS further extends Photo-SLAM by incorporating splat-wise back-propagation significantly improving GS map rendering speed and visual quality~\cite{mallick_taming_2024, feng_cartgs_2024}.

\section{System Overview}
\label{sec:methodology}

\begin{figure*}
    \centering
    \includegraphics[width=\textwidth]{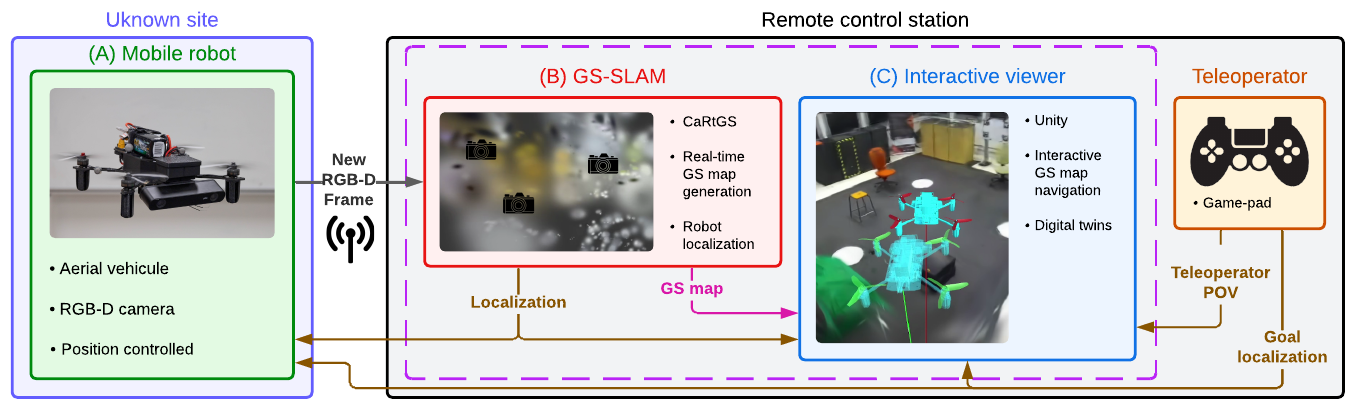}
    \caption{A mobile robot in the unknown site (A) streams RGB-D images to the remote control station, where a GS-SLAM algorithm (B) generates a 3D map in real-time and localizes the robot within it. This GS map and localization information are visualized in an interactive viewer (C), where a goal proxy replicates the teleoperator's game-pad inputs in real-time, guiding the robot's digital twin for precise tele-exploration. The brown arrows indicate the data exchanges that are managed using ROS~2. The pink arrow materializes the efficient GPU-based integration between GS-SLAM and the interactive viewer.}
    \label{fig:Pipeline_overview}
\end{figure*}

Our remote teleoperation system is based on 3 components presented in Fig.~\ref{fig:Pipeline_overview}): (A) a mobile robot, streaming RGB-D images; (B) a state-of-the-art GS-SLAM algorithm that processes the RGB-D images to generate in real-time a photorealistic 3D map of the site and a robust localization; and (C) an interactive viewer, that presents the photorealistic view of the robot with alongside relevant information for precise remote teleoperation. Our mobile robot is an aerial vehicle, requiring precise localization and fast response to maintain stability and maneuverability, unlike less dynamic ground-based exploration robots.

\begin{figure}[h]
    \centering
    \includegraphics[width=\columnwidth]{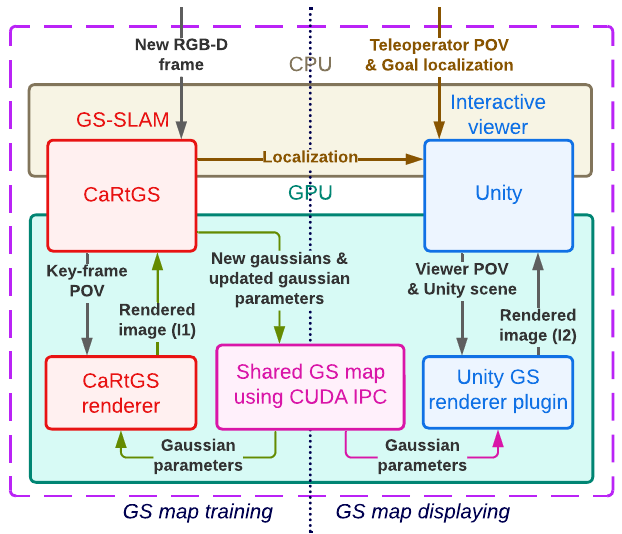}
    \caption{This diagram illustrates how CaRtGS and Unity connect. Each component runs in a separate process and has its own GS renderer. The GS map is stored on the GPU and shared between the two processes via CUDA IPC (Inter-Process Communication). By directly accessing the GS map, Unity can update its photorealistic 3D scene with minimal latency, while CaRtGS performance remains unaffected. This figure uses the same color code as Fig.~\ref{fig:GS_SLAM} and Fig.~\ref{fig:Pipeline_overview} to represent identical components.}
    \label{fig:CUDA_IPC}
\end{figure}

Following the discussion of Sec.~\ref{sec:related_works}, our GS-SLAM algorithm is based on CaRtGS\footnote{\url{https://github.com/DapengFeng/cartgs}}, due to its superior rendering quality, speed, precise localization and support for loop closure. A key challenge is that views (I1) and (I2) in Fig.~\ref{fig:GS_SLAM} from the GS map must be generated simultaneously for both the optimizer updating Gaussian parameters and the interactive viewer for the teleoperator. This is typically done by alternating scene rendering between the two, which causes interferences and trade-offs between map generation speed and interactive viewer frame rate, as shown in Fig.~\ref{fig:GS_SLAM}. To address this issue, we have modified CaRtGS to separate and decouple these two processes as illustrated in Fig.~\ref{fig:CUDA_IPC}. Each process now has its dedicated GS renderer, utilizing CUDA IPC (Inter-Process Communication) to enable simultaneous access to the GS map by both the optimizer and the interactive viewer, directly on the GPU. Our approach is modular by design, enabling straightforward integration of other GS-SLAM algorithms beyond CaRtGS.

\begin{figure}[h]
   \centering
   \includegraphics[width=\linewidth]{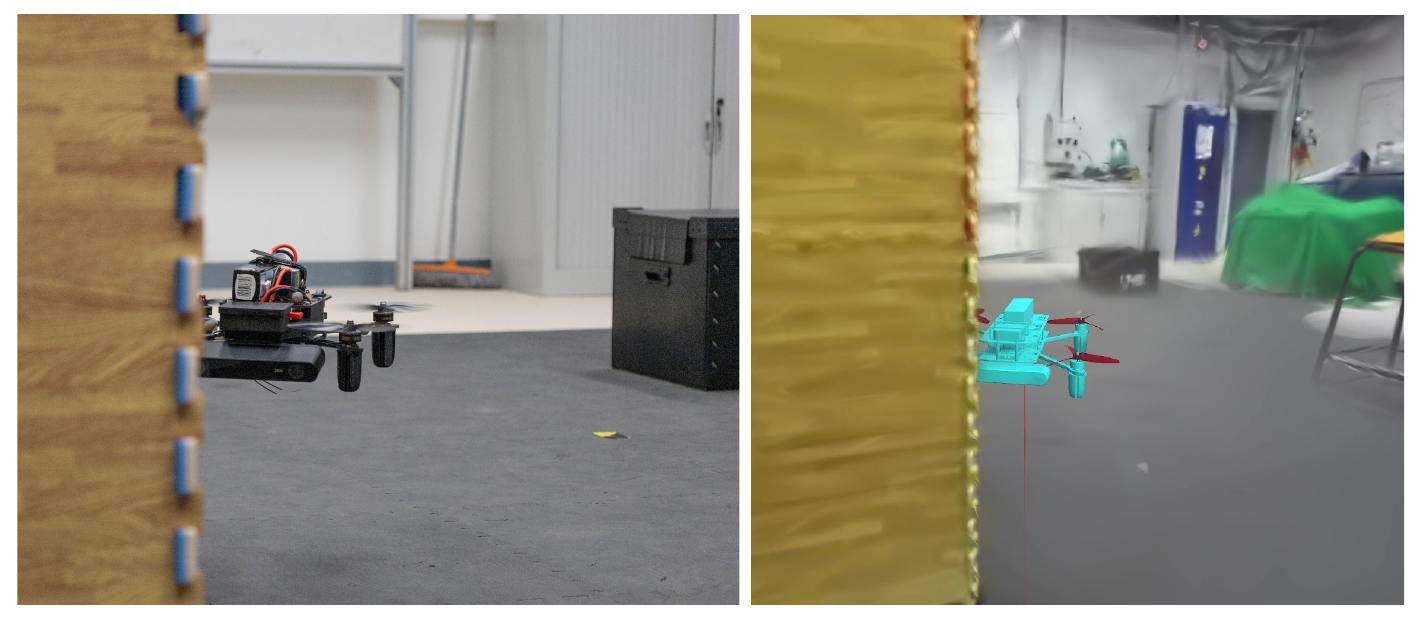}
   \caption{Occlusion between the robot's digital twin and the GS map.}
   \label{fig:scene_occlusion}
\end{figure}

\begin{figure*}[h]
    \centering
    \includegraphics[width=0.8\textwidth]{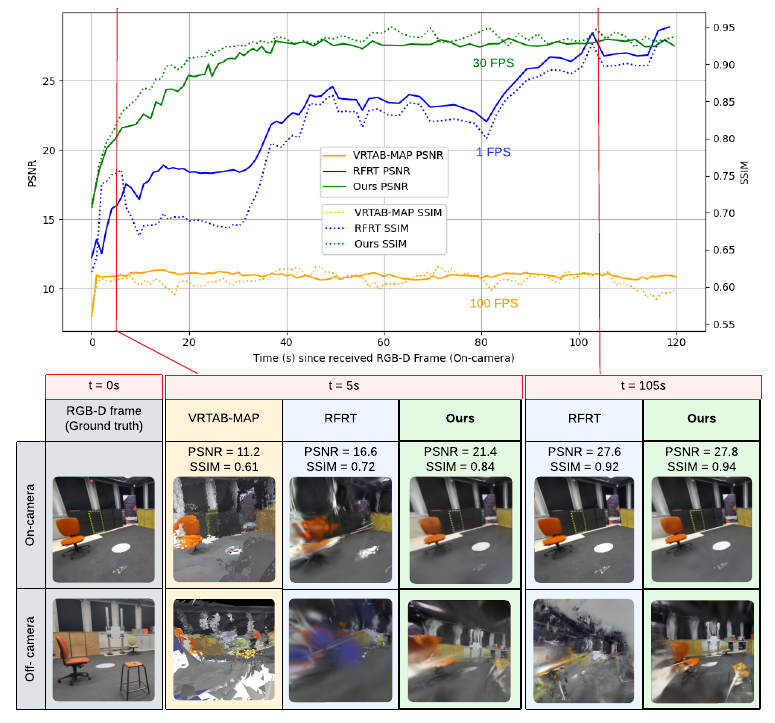}
    \caption{Performance evaluation of the proposed system and its comparison with state-of-the-art online map-based teleoperation systems.}
    \label{fig:Benchmark}
\end{figure*}

Our interactive viewer is based on Unity, and leverages a customized GS rendering plugin\footnote{\url{https://github.com/clarte53/GaussianSplattingVRViewerUnity}}. This plugin can render multiple viewpoints simultaneously, which is necessary for VR headsets. The system is controlled using a game-pad shown in Fig.~\ref{fig:Overview}, offering an intuitive interface inspired by third-person video games. One joystick is dedicated to controlling the robot's goal position in the scene, while the other is dedicated to managing the camera view, allowing the operator to explore the map from different angles. In this setup, the camera is always focused on the robot, to help with its teleoperation. This dual-control system lets the operator manage independently the robot's motion and the virtual camera's view. To facilitate precise teleoperation, the interactive viewer shown in Fig.~\ref{fig:Unity} integrates digital twins of the robot's current and goal positions in the GS scene. This integration in Unity allows also representing accurate occlusions, which are essential for correct situational awareness as shown in Fig.~\ref{fig:scene_occlusion}.

\section{System performance}
\label{sec:teleop_performances}

The performances of our system are evaluated in the following on a laptop equipped with an Intel i7 vPRO CPU, 64GB of RAM and an NVIDIA RTX Ada 4000 GPU. The image resolution used is 672x376. The primary interface for the teleoperator being the interactive viewer, we focus on the following three performance metrics:

\begin{itemize}
    \setlength{\itemsep}{5pt}
    \item Map visual accuracy, to provide reliable spatial information without distortion and misinterpretation, evaluated using Peak Signal-to-Noise Ratio (PSNR) and Structural Similarity Index (SSIM).
    
    \item Map generation speed, to enable fast and reactive exploration, measured as the time taken for each incoming RGB-D frame to reach its final PSNR and SSIM values. 
    
    \item Interactive viewer responsiveness, to enable fluid interactions with minimum lag, measured in Frames Per Second (FPS).
\end{itemize}

Our current evaluation is restricted to views that were observed during training. Investigating quantitative performance from unseen, off‑trajectory viewpoints is a promising avenue for future work, especially since existing GS-SLAM benchmarks do not yet address this scenario. We evaluate a set of sampled key-frames from a dedicated dataset recorded during a typical exploration scenario to compare three state-of-the-art online map-based teleoperation systems: VRTAB-MAP~\cite{stedman_vrtab-map_2022}, the Radiance Field for Robotic Teleoperation (RFRT) method described in~\cite{wilder-smith_radiance_2024}, and our system.

First of all, our rendering architecture based on CUDA IPC (Fig.~\ref{fig:CUDA_IPC}) allows a fluid and responsive experience with 30 FPS in the interactive viewer without compromising map generation speed, in contrast with RFRT which only reaches 1 FPS on our hardware. While maintaining comparable image quality for previously seen views and improving it for unseen views, our system generates maps 3 to 5 times faster than RFRT, depending on the desired image quality as shown in Fig.~\ref{fig:Benchmark}.  Our system matches the map generation speed of VRTAB-MAP while delivering significantly superior image quality in all situations. Combining the map generation speed of VRTAB-MAP with the rendering quality of RFRT and a fluid, responsive interface, our system makes no compromise delivering real-time photorealistic maps for increased SA and efficient teleoperation, as evaluated next.

\section{System Validation}
\label{sec:experiments}

\begin{figure}[h]
     \centering
     \includegraphics[width=\linewidth]{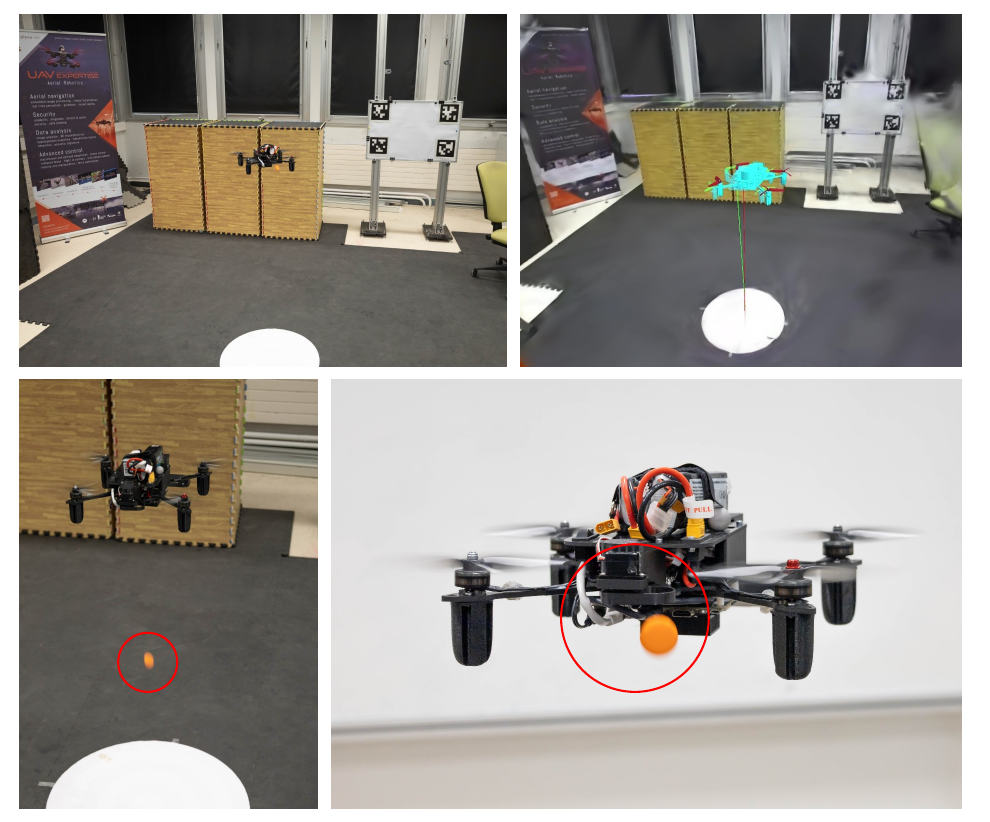}
     \caption{This figure illustrates the experimental setup, highlighting various components of the teleoperation system. The top-left image shows the real-world environment, while the top-right image presents the interactive viewer's perspective. The bottom-left image depicts the moment the marker is dropped onto the target, and the bottom-right image offers a close-up view of the marker-dropping mechanism mounted on the drone. In the interactive viewer, lasers projected from the quadcopter aid the teleoperator in assessing the marker's impact position on the target. Video of the experiments is available at: \url{https://www.youtube.com/watch?v=-Md49rKkV8I}}
     \label{fig:Dropping}
 \end{figure}

We use a custom-built quadcopter shown in Fig.~\ref{fig:Unity} with standard onboard architecture and a PX4 Autopilot to handle low-level controls and sensor fusion along with protocol management and security. The quadcopter is programmed to continuously follow the target position set by the operator using localization from CaRtGS, with a delay dictated by its own dynamics and CaRtGS's ORB-SLAM~3 backend processing time. It is equipped with a Stereolabs ZED 2i RGB-D camera mounted with a 20\textdegree\mbox{} tilt downward, ensuring a more balanced view of the scene. RGB-D images are compressed onboard using an Nvidia Jetson Xavier running the GPU-based ZED SDK, and streamed via WiFi to the ground station for further processing by the GS-SLAM algorithm. Communication is managed through ROS2 (brown arrows in Fig.~\ref{fig:Pipeline_overview}) and WiFi-based Mavlink. The quadcopter is also equipped with a marker-dropping mechanism activated by the operator and shown in Fig.~\ref{fig:Dropping}.

\begin{figure}[h]
    \centering
    \includegraphics[width=\columnwidth]{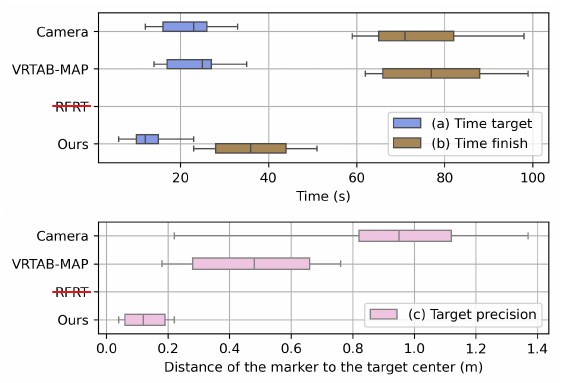}
    \caption{Real-world performance evaluation of the proposed system. Each box plot displays the median (central line), inter-quartile range (box edges from Q1 to Q3) and whiskers showing data spread.}
    \label{fig:Experiment}
\end{figure}

To assess key aspects of situational awareness outlined in the introduction~\ref{Sec:Introduction}, five human operators with no prior experience in quadcopter piloting participate in the evaluation. They are tasked to explore an unknown flight arena and deploy markers on designated targets by remotely teleoperating the quadcopter using multiple methods. Each teleoperator first spends five minutes using the game-pad and the interactive viewer to learn the control interface. Then, he has one minute after takeoff to explore the flight arena with the quadcopter before receiving an instruction to drop a marker on one of the three identical white targets placed on the ground. The selected target is identified by contextual cues from its surroundings (for example, “the target next to the red chair”), requiring the remote operator to interpret the scene and understand the objects present at the site. Finally, to confirm successful self-localization and effective map usage, the operator is asked to navigate back to the original starting position. Participants are instructed to perform this task multiple times, first relying only on direct camera video feed, then using VRTAB-MAP, RFRT and our system. The time taken to reach the target (a), the time to return to the starting position (b), and the impact precision on the target (c) are presented in Fig.~\ref{fig:Experiment}. 
 
RFRT couldn't be used by any teleoperator because of its insufficient Rviz-based interactive viewer frame-rate (1 FPS). The direct camera feed provides real-time visibility, allowing operators to complete the task. However, it lacks spatial context, resulting in poor marker placement accuracy. VRTAB-Map improves targeting accuracy by offering multiple viewpoints, but 2 out of 5 operators failed to accurately identify the target due to poor visual map quality. Our system outperforms all other methods, achieving both the fastest completion times and the highest placement precision. Additionally, the smaller box plot variance indicates greater consistency, reducing the impact of random variations in performance and ensuring a more reliable exploration experience. This translates directly into better situational awareness for the operators, enabling more efficient decision-making and control.

The ideal scenario is for the mapping system to update the scene as fast as the robot navigates its environment, so the teleoperator doesn’t experience any delay waiting for the map to generate. When the robot is moving straight, the camera captures distant details, but during turns, new, previously unseen areas appear rapidly. With a wide field of view, the system already sees parts of the scene during a turn and can begin mapping immediately. We experimentally found out that a minimum PSNR of 20 is necessary for clear scene comprehension during teleoperation. For instance, in our aerial vehicle teleoperation case where the robot rotates 90 degrees in about 4 seconds, our method produces a map with a PSNR of 21 in just 5 seconds as shown in Fig.\ref{fig:Benchmark}, nearly matching the robot’s movement speed, whereas the RFRT method requires 35 seconds to reach the same quality.

\section{Conclusion}

In this paper, we presented a remote teleoperation system that leverages recent advancement in GS-SLAM to generate photorealistic 3D maps of unknown environments in real-time. By decoupling the scene-optimization and rendering processes through GPU-based inter-process communication, our approach achieves high visual accuracy, low-latency visualization, and a responsive user interface. A comparative study further confirms its superiority over state-of-the-art solutions, effectively resolving the longstanding trade-off in online map-based teleoperation systems. 

Experimental results demonstrate that our system supports teleoperators in making rapid, well-informed decisions under limited prior knowledge conditions. User studies on scene interaction and exploration in an unknown site via quadcopter teleoperation showed that participants using our system completed missions faster and more accurately than with existing methods. These findings strongly suggest that the combination of GS-SLAM and a high-performing interactive viewer effectively elevates situational awareness and mission reliability in an unknown-site teleoperation scenario.

The proposed approach lays an effective foundation for the next generation of teleoperation systems, empowering operators to handle increasingly complex tasks and domains with confidence. Future work may extend this work to dynamic or partially changing environments, multi-robot settings, and more advanced human–robot collaboration scenarios.

\section*{Acknowledgment}
\footnotesize{Research and equipment has been partially supported by ROBOTEX 2.0 (Grants ROBOTEX ANR-10-EQPX-44-01 and TIRREX ANR-21-ESRE-0015) and Framatome.}

\bibliographystyle{IEEEtran} 
\bibliography{references} 

\end{document}